\documentclass[nohyperref]{article}

\usepackage[switch]{lineno}
\usepackage{microtype}
\usepackage{graphicx}
\usepackage{subfigure}
\usepackage{booktabs} % for professional tables

\usepackage{hyperref}

\usepackage[accepted]{icml2023}

\usepackage{amsmath}
\usepackage{amssymb}
\usepackage{mathtools}
\usepackage{amsthm}
\usepackage{widetext}

\usepackage[capitalize,noabbrev]{cleveref}

\theoremstyle{plain}

\theoremstyle{definition}

\theoremstyle{remark}

\usepackage[textsize=tiny]{todonotes}

\icmltitlerunning{Neural stochastic models for collective movement}

\begin{document}

\twocolumn[
\icmltitle{Discovering mesoscopic descriptions of collective movement \\
with neural stochastic modelling}



\icmlsetsymbol{equal}{*}

\begin{icmlauthorlist}
\icmlauthor{Utkarsh Pratiush}{uot,ece}
\icmlauthor{Arshed Nabeel}{ces,imi}
\icmlauthor{Vishwesha Guttal}{ces}
\icmlauthor{Prathosh AP}{ece}
% \icmlauthor{Firstname1 Lastname1}{equal,yyy}
% \icmlauthor{Firstname2 Lastname2}{equal,yyy,comp}
% \icmlauthor{Firstname3 Lastname3}{comp}
\end{icmlauthorlist}

% \icmlaffiliation{sire}{School of Interdisciplinary Research, Indian Institute of Technology, Delhi}
\icmlaffiliation{uot}{University of Tennesse, Knoxville, United States.}
\icmlaffiliation{ces}{Center for Ecological Sciences, Indian Institute of Science, Bengaluru, India.}
\icmlaffiliation{imi}{IISc Mathematics Initiative, Indian Institute of Science, Bengaluru, India.}
\icmlaffiliation{ece}{Deparment of Electrical Communication Engineering, Indian Institute of Science, Bengaluru, India.}

\icmlcorrespondingauthor{Vishwesha Guttal}{guttal@iisc.ac.in}
\icmlcorrespondingauthor{Prathosh AP}{prathosh@iisc.ac.in}

\icmlkeywords{Machine Learning, ICML}

\vskip 0.3in
]



\printAffiliationsAndNotice{}  % leave blank if no need to mention equal contribution

\begin{abstract}

Collective motion is an ubiquitous phenomenon in nature, inspiring engineers, physicists and mathematicians to develop mathematical models and bio-inspired designs. Collective motion at small to medium group sizes ($\sim$10-1000 individuals, also called the \emph{`mesoscale'}), can show nontrivial features due to stochasticity. Therefore, characterizing both the deterministic and stochastic aspects of the dynamics is crucial in the study of mesoscale collective phenomena. Here, we use a physics-inspired, neural-network based approach to characterize the stochastic group dynamics of interacting individuals, through a \emph{stochastic differential equation (SDE)} that governs the collective dynamics of the group. We apply this technique on both synthetic and real-world datasets, and identify the deterministic and stochastic aspects of the dynamics using \emph{drift} and \emph{diffusion} fields, enabling us to make novel inferences about the nature of order in these systems. 

% Discovering SDE's is an important step towards studying collective behaviour in ecological systems. Methods such as equation learning helps in it but the polynomial fitting aspect to it imposes certain biases in modelling. In this work we propose a neural network methods which are equation free thus abstains from assuming any biases. Through this work we are able to show asymmetry which can be seen in data to be seen in learnt drift and diffusion functions which is not caught by equation learning methods. We also extend the work where we make the drift and diffusion of SDE an extra function of position which we argue can be achieved only using our method for discovering SDE's.
\end{abstract}

\newcommand{\bx}{\mathbf{x}}
\newcommand{\bv}{\mathbf{v}}
\newcommand{\bm}{\mathbf{m}}
\newcommand{\br}{\mathbf{r}}
\newcommand{\bof}{\mathbf{f}}
\newcommand{\bg}{\mathbf{g}}
\newcommand{\boldeta}{\boldsymbol{\eta}}
\newcommand{\pfg}{p_{\bof, \bg}}
\newcommand{\modm}{|\bm|}
\newcommand{\trel}{T_{\text{rel}}}

% \linenumbers

\section{Introduction}

Collective motion is a phenomenon that is observed in natural and synthetic systems, and has fascinated physicists and biologists alike~\cite{vicsek2012collective,sumpter2010collective,camazine2020selforg}. Many systems across scales---such as microscopic organisms~\cite{dinet2021linking,be2019statistical}, cells~\cite{rorth2009cellmigration, alert2020physical}, human crowds~\cite{chen2018social}, and synthetic active matter~\cite{ramaswamy2010mechanics,ramaswamy2017active}---exhibit self-organized collective movement. How the seemingly simple behaviour and interactions of the individuals give rise to the complex self-organised emergent dynamics of the group, is one of the central questions in the study of collective dynamics.

At the level of individual organisms, animal behaviour is complex, and modelling every aspect of individual {\it stochastic} animal behaviour seems an unattainable goal. While the stochastic effects typically average out in the limit of infinite (or sufficiently large) group sizes, real-world animal groups are finite, and often small to medium sized (10 to 1000 individuals). At these \emph{`mesoscopic'} scales, the individual-level stochasticity can affect on the group dynamics in non-trivial ways. Therefore, a correct description of the dynamics at the mesoscale should incorporate stochasticity~\cite{mckane2004stochastic, yates2009locust, biancalani2014prl, bruckner2019stochastic, jhawar2020fish}. 

Owing to the complexity at the individual behavioural level, one often uses a coarse-grained dynamical description of the collective system. A physicist's way of approaching this this is to quantify the state of the group, the so called \emph{order parameter}, whose dynamics is then modelled to obtain a parsimonious description of the system. In the context of collective motion, order parameter can be the \emph{group polarisation}---a quantity analogous to the magnetisation of spin systems~\cite{vicsek1995novel}. To study the dynamics of order parameter at the mesoscale, the inherent stochasticity becomes crucial; and the order parameter dynamics must be modelled as a stochastic process. A commonly used framework for continuous stochastic processes such as this is the \emph{diffusion process}, which can be described using a \emph{stochastic differential equation (SDE)}. An SDE describes a stochastic process by disentangling it into a deterministic term, called the \emph{drift}, and a stochastic term, called the \emph{diffusion}~\cite{van1992stochastic, gardiner2009}.

Analytically deriving coarse-grained SDE descriptions starting from individual-level interactions is an arduous task, even for highly simplified toy models of collective behaviour~\cite{mckane2004stochastic,biancalani2014prl,van1992stochastic,jhawar2019bookchapter}. Indeed, for real-world collective systems, one seldom has a detailed understanding of the individual behaviour; therefore, obtaining a group-level description seems hopeless. 

We address this challenge by tackling the inverse problem: we use a data-driven approach to estimate mesoscopic SDEs directly from observed group trajectory data. From the individual movement trajectories, we compute the \emph{polarization order parameter}, quantifying the level of alignment in the group. We then use a physics-inspired, neural-network based approach~\cite{dietrich2021esde,evangelou2022esde} to fit an SDE model to describe the temporal dynamics of polarization. To facilitate the interpretability of the discovered neural SDE---something which is notoriously hard in neural-network-based approaches---we propose a method to visualize the discovered drift and diffusion as fields. The drift and diffusion fields enable us to readily identify the deterministic equilibria, structure of stochasticity, etc.

We use this approach to study both simulated as well as real-world fish (species \emph{Etroplus suratensis}) schools. For a well-studied simulation model of collective behaviour~\cite{buhl2006disorder,biancalani2014prl,dyson2015onset,jhawar2020fish,jhawar2020inferring}, the neural SDE models can accurately recover the mean field SDEs---and hence identify underlying interactions. We also discover SDEs from a real world dataset of schooling fish, where we observe the signatures of \emph{noise-induced order}, a counter-intuitive phenomenon where intrinsic noise serves to increase group-level order rather than destroy it~\cite{biancalani2014prl,jhawar2020inferring}. This was first observed by~\cite{jhawar2020fish}, who used the conventional SDE discovery methods~\cite{jhawar2020inferring,tabar2019book}. Our neural-network-based approach recapitulates the same result; we emphasize that this result is non-obvious, since neural SDEs capture a much wider hypothesis class than the conventional SDEs.

The main contributions of this study are as follows:
\begin{itemize}
    \item We demonstrate the efficacy of neural-network-based approaches~\cite{dietrich2021esde,evangelou2022esde} for discovering mesoscale stochastic dynamics for collective behaviour data, using both simulated datasets where the mesoscale SDEs are analytically established.
    \item To make the results of the \emph{grey-box} neural network techniques more interpretable, we propose novel visualization techniques to visualize the drift and diffusion fields in an easily interpretable form.
    \item We apply these techniques to study the mesoscale dynamics in real-world experimental datasets of fish schools~\cite{jhawar2020fish}, thereby demonstrating the significance of these techniques in animal behaviour and ecology. To the best of our knowledge, ours is the first study that applies these techniques to a real world dataset of animal collective behaviour, or even the broader field of ecology.
\end{itemize}

\paragraph{Related work:} A class of techniques for estimating SDEs from data involve computing the \emph{jump moments} from the observed time-series, and estimating the drift and diffusion coefficients of the SDE using certain conditional averages of the jump-moments~\cite{tabar2019book, gardiner2009}. However, these conditional averages are often noisy and inaccurate, especially when the available data is limited. More recently, approaches have been developed to use jump moments to obtain smooth estimates of the drift and diffusion coefficients, using kernel density estimation, sparse regression  etc.~\cite{gorjao2019kramersmoyal,nabeel2022pydaddy,boninsegna2018sparse,wang2022data}. Another approach is to use parametric estimation techniques, by assuming an appropriate parametric form for the drift and diffusion functions~\cite{nielsen2000parameter}---however, these techniques have the drawback of having to know the parametric form of the SDE \emph{a priori}. 

The neural-network-based approach used in this study was first introduced by~\cite{dietrich2021esde}, and was subsequently applied to derive effective SDE descriptions from simulated Brownian dynamics~\cite{evangelou2022esde}. This is a \emph{`physics-inspired, grey-box'} approach, where partial knowledge and assumptions about the underlying physics is used to make the neural network model at least partly interpretable. 

Alternative approaches for data-driven neural SDE modelling also exist in literature, such as~\cite{song2020score, li2020scalable}, which build on previous work on neural ODEs~\cite{chen2018neural, liu2020rode}. Alternative approaches work by matching moments~\cite{o2021stochastic} or ensemble distributions~\cite{o2021stochastic}, or use VAEs to learn an SDE in a latent space~\cite{hasan2021identifying}. We chose the approach of~\cite{dietrich2021esde} for its simplicity and flexibility because of the way the drift and diffusion are explicitly modelled and incorporated into a straightforward loss-function.

\section{Background \& Methods}

\begin{figure*}
    \centering
    \includegraphics[width=\textwidth]{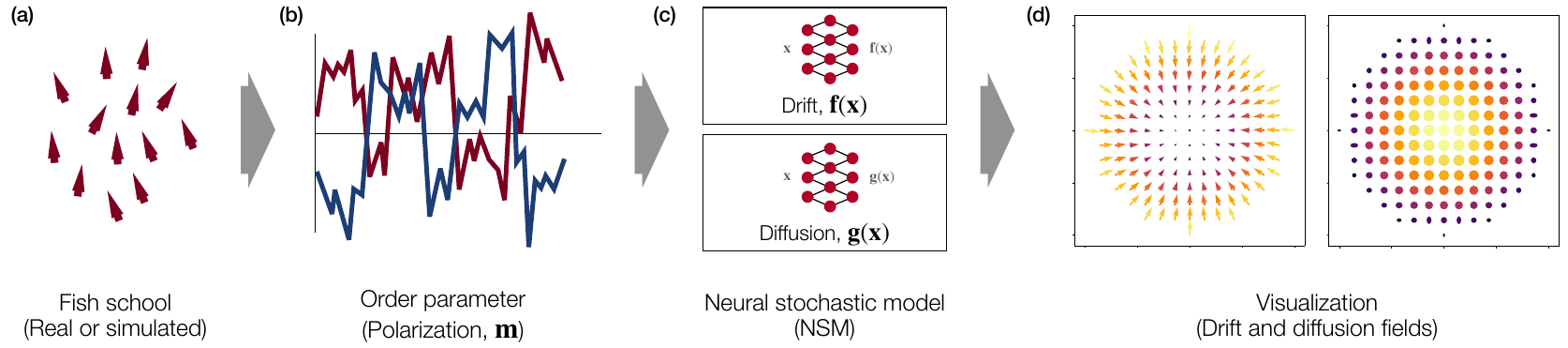}
    \caption{\textbf{Schematic of the estimation procedure.} (a) Individual trajectories are recorded for a real or simulated collective motion. (b) From the individual trajectories, the polarization order parameter ($\bm$) is computed, quantifying the degree of alignment in the school. (c) Neural networks are trained, using a likelihood framework, to estimate the drift ($\bof$) and diffusion ($\bg$) functions. (d) For interpretation, the drift and diffusion functions are visualized as vector and tensor fields respectively. This figure shows a schematic illustration, see Section~\ref{sec:visualization} and Fig.~\ref{fig:fields-sim}, \ref{fig:fields-real}.}
    \label{fig:schematics}
\end{figure*}

% This can go to the appendix.
% \subsection{Experimental setup}
% We use a dataset of schooling fish in this study. Groups of \textit{Etroplus suratensis} fish were placed in a shallow circular tank (so that their movement is restricted to two dimensions), and their movement was recorded using a high-resolution camera. From the recorded videos, we tracked the positions $\bx_i(t)$ and velocities $\bv_i(t)$ of each fish $i$ using a tracking software \emph{trex} (CITE). From the individual fish trajectories, the \emph{group polarisation order parameter} was computed as the average of the normalized velocity vectors, i.e. $\bm = \frac{1}{N} \sum_i \hat \bv_i(t)$. The goal is to model the stochastic dynamics of $\bm$ using an SDE.

\subsection{\label{sec:datasets} Datasets}

We use both simulated as well as real-world datasets of collective movement in this study. The simulated data comes from an idealized model of collective movement, for which the mesoscopic dynamics are well-studied~\cite{jhawar2020fish,jhawar2019bookchapter}. The real-world dataset is an open-access dataset (\cite{jiteshjhawar_2020_3632470}), consisting of trajectories of fish schools of different group sizes.

\paragraph{Agent based models for collective movement.} We use simulations from an idealized agent based model of collective movement, with $N = 30$ individuals. Each individual is described solely by its velocity vector $\bv_i$. The magnitude of $\bv_i$ is assumed to be constant across all agents and across time, only the direction changes. Direction changes can happen asynchronously at random time-points, in one of two ways:

\begin{enumerate}
    \item An agent $i$ can spontaneously turn and choose a new direction, i.e. $\theta_i(t) \leftarrow \eta$ for $\eta \sim \mathsf{Unif}[-\pi, +\pi)$.
    \item An agent $i$ can copy the direction of one or more other agents, chosen randomly from the entire group.
\end{enumerate}

The turn and copy events themselves happen as a Poisson process. The models can be classified as \emph{pairwise} or \emph{ternary} interaction models based on how many neighbours are copied. The mesoscopic SDEs for these models are already known in literature~\cite{jhawar2020fish,jhawar2019bookchapter}---also see Appendix~\ref{sec:analytical}. Therefore, the model simulations act as a test-bed to verify the efficacy of the data-driven estimation procedure.

\paragraph{Real-world datasets of schooling fish.} 
We used an openly available dataset of fish schooling~\cite{jhawar2020fish,jiteshjhawar_2020_3632470}. This dataset consists of the position $\bx_i(t)$ and velocity $\bv_i(t)$ trajectories of fish (species \emph{Etroplus suratensis}), swimming in a circular arena. From the individual fish trajectories, the polarization order parameter is computed as detailed in the following section. We used datasets with $N = 15, 30$ and $60$ fish. The data was available at a sampling interval of $0.12s$. 
% Removing detailed description of dataset for anonymity.
% The fish were placed in a shallow, circular tank (radius 30cm) so that their motion was restricted to two dimensions. The movement of the fish was recorded using a high-resolution video camera, for a duration of ~1 hr. From the videos, the trajectories of the individual fish were extracted. From the individual fish trajectories, the polarization order parameter is computed as detailed in the following section. 

\subsection{\label{sec:ordparam} Mesoscopic descriptions of collective dynamics}

While modelling collective movement, the emergent dynamics of the group can often be characterized in terms of an \emph{order parameter}, which characterizes the degree of order in the group. An oft-used order parameter is the group polarization, which captures the level of alignment among the individuals of the group.

For a group of $N$ individuals, each with positions $\bx_i$ and velocities $\bv_i$, the polarization can be computed as the mean of the normalized velocity vectors, i.e.,
\begin{align}
    \bm = \frac{1}{N} \sum_{i=1}^{N} \frac{\bv_i}{|\bv_i|}
\end{align}

Modelling the time-evolution of $\bm$ allows us to gain an understanding of the overall emergent dynamics of the group. For relatively small group sizes, idiosyncrasies in the individual $\bv_i$'s can have a significant effect on the group dynamics. These result in stochastic fluctuations which cannot be ignored while modelling the dynamics of $\bm$. Therefore, we use the framework of \emph{stochastic differential equations (SDE)} to model the $\bm$, which models both deterministic as well as stochastic aspects of the time-evolution of $\bm$.

\subsection{Data-driven discovery of stochastic differential equations}

Given the time series of the 2-dimensional polarization vector $\bm(t)$ sampled with a finite time-interval $\Delta t$, 
our goal is to discover an It\^{o} stochastic differential equation (SDE) model that explains the time-series, of the following form:

\begin{align}
    \frac{d \bm}{dt} = \bof(\bm) + \bg(\bm) \cdot \boldeta(t)  \label{eq:sde}
\end{align}

where $\boldeta$ is 2-dimensional white noise process. Here, $\bof$ is called the \emph{drift} function and $\bg$ is called the \emph{diffusion} function. The goal in the SDE discovery problem is to discover representations for $\bof$ and $\bg$.
% , either in the form of an analytical expression~\cite{brunton2016sindy,boninsegna2018sparse,nabeel2022pydaddy}, or based so-called \emph{equation-free} approaches.
In this work, 
% we take the latter approach, where the drift and diffusion functions are approximated using neural networks. We 
we use a likelihood-based framework, first introduced by~\cite{dietrich2021esde}. Here, $\bof$ and $\bg$ are represented using neural networks, trained to maximize a likelihood function based on the finite-time transition probability. 

Specifically, let $p(\cdot, t_1 | \bm_0, t_0)$ be the probability density function of $\bm(t_1)$ conditional to $\bm(t_0) = \bm_0$. Then, for a small time-step $\Delta t$, $p(\cdot, t + \Delta t | \bm_0, t)$ can be approximated based on the stochastic Euler approximation (also known as the Euler Maruyama approximation) as:

\begin{align}
    p(\cdot, t + \Delta t | \bm_0, t) \sim \mathcal{N}\left( \bm_0 + \bof(\bm_0) \cdot \Delta t, \bg(\bm_0) \cdot \Delta t \right)
\end{align}

Parameterizing $p$ in terms of the drift and diffusion functions as $\pfg$, this gives rise to the following log-likelihood loss function which can be maximized to fit $\bof$ and $\bg$:

% \begin{widetext}
\begin{align}
    &\mathcal{L}\left(\bof, \bg \middle| \bm_0, \bm_1 \right) 
        = \log \pfg \left(\bm_1, t + \Delta t \middle | \bm_0, t \right) && \nonumber \\
    &    = \frac{(\bm_1 - \bm_0 - \bof(\bm) \Delta t)^2}{\bg(\bm_0)^2 \Delta t} +
           \log \left | \bg(\bm_0)^2 \Delta t\right | + \log 2 \pi \label{eq:likelihood}
\end{align}
% \end{widetext}
where $\bm_0$ and $\bm_1$ are successive points in the time series dataset, sampled $\Delta t$ apart. The loss function $\mathcal L$ can be minimized to fit an appropriate representation of $\bof$ and $\bg$.

We represent $\bof$ and $\bg$ using neural networks, which were trained to minimize the loss function in Equation~\ref{eq:likelihood}--see Section~\ref{sec:implementation} for further details about the implementation. Figure~\ref{fig:schematics} illustrates the full data-driven model discovery procedure, starting from the individual trajectories, computing the polarization order parameter, estimating drift and diffusion using neural networks, and visualizing them as fields.

\subsection{\label{sec:implementation} Implementation details}

\paragraph{Dataset preparation:} The agent-based schooling model was implemented and simulated in MATLAB, using the stochastic simulation algorithm (SSA)~\cite{gillespie2007stochastic}. From the trajectories of individual velocity vectors produced by the simulation, the group polarization $\bm(t)$ was computed (Section~\ref{sec:ordparam}), which was used for training the model. Similarly, for the fish schooling dataset, the tracked individual trajectories (see Section~\ref{sec:datasets}) were used to compute the group polarization $\bm(t)$.

In both the agent-based models as well as real-world datasets, there is rotational and mirror symmetry in the dynamics (the simulations has no angular bias, while the fish schools were swimming in a circular arena). Therefore, we expect these symmetries to carry over to the mean-field models as well. To encourage the neural network to learn these symmetries, we augment the time-series of $\bm(t)$ with rotated (16 rotations equally spaced between $-\pi$ and $\pi$) and flipped (horizontal and vertical flips) versions of itself. 

\paragraph{Architecture and training details:} The drift and diffusion functions $\bof$ and $\bg$ were represented using fully-connected neural networks with 5 layers and 150 neurons per layer, with ELU activation function~\cite{clevert2015elu}. The network was trained with a batch size 512 and test-train split of 90:10, with the Adamax optimizer. 

Once the models are trained, to validate the discovered models, we generate simulated trajectories using the model, and compare the probability density (histogram) and autocorrelation function of the simulated trajectory to that of the input trajectory. The similarity between the histograms was quantified using the Wasserstein metric, $W_1$~\cite{ramdas2017wasserstein}. The similarity of the autocorrelation was quantified using the relative timescale discrepancy, $\trel = | \tau - \hat \tau | / \tau$, where $\tau$ and $\hat \tau$ are the autocorrelation times of the real and simulated time series respectively. A good fit according to these metrics ensures that the discovered dynamical model matches the actual dynamics at least in the weak (distributional) sense~\cite{kloeden1992stochastic}.
% We carried out all our model training in Tensorflow[cite]. A simple feed-forward neural network architecture with 5 layers was used, each layer having 100 perceptrons. We used elu activation function[cite], which was decided based on trial and error. Training was carried on with batch size of 128. Ten percent of data was set aside for testing. Adamax optimizer module[cite] was used for optimising the network.

% Two networks were initialized, one predicting the mean and other predicting the variance. Both the neural networks were optimised simultaneously using the combined loss function.

% \paragraph{Modelling non-autonomous dynamics with driver variables.} Consider a situation where the dynamics is also influenced by additional driver variable(s) $\br$, which are not part of the dynamical variables. The dynamics of $\bm$, the dynamical variable, now obeys the following equation:

% \begin{align}
%     \frac{d \bm}{dt} = \bof(\bm, \br) + \bg(\bm, \br) \cdot \boldeta(t)  \label{eq:sde-2}
% \end{align}

% With the likelihood framework, extending the estimation procedure to this scenario becomes straightforward---simply replacing $\bof$ and $\bg$ in Eq.~\ref{eq:likelihood} with the modified functions allows one to estimate a stochastic dynamical model with driver variables. Here, we use this modification to  study how the polarisation dynamics of the fish school depends on the proximity the tank boundary, by introducing $\br$, the distance from from the tank center, as a driver variable.

\begin{figure*}
    \centering
    \includegraphics[width=\textwidth]{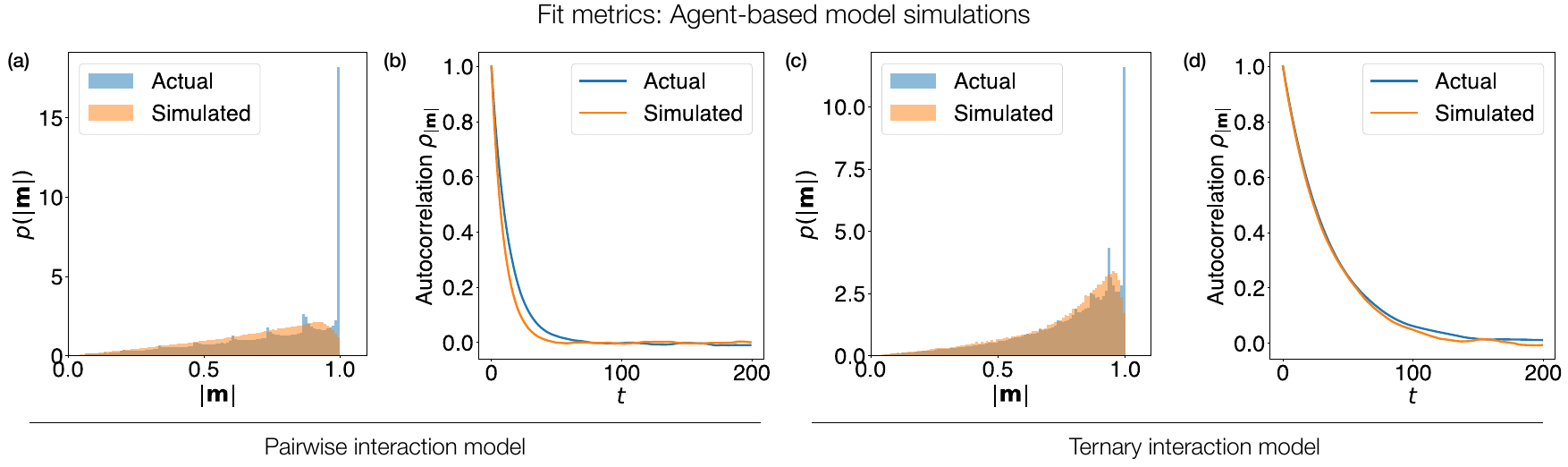}
    \caption{\textbf{Goodness-of-fit metrics for mesoscale models for agent-based simulations.} (a, c) Comparison of histograms of $|\bm|$ from the actual time series as well as a time series simulated with the discovered neural stochastic model; for the pairwise (a) and ternary (c) interaction models. (b, d) A similar comparison between the autocorrelation functions of the actual and simulated $|\bm|$ time series from the two models.}
    \label{fig:metrics-sim}
\end{figure*}

\subsection{\label{sec:visualization}
Visualizing the drift and diffusion fields}

Since the drift and diffusion functions are represented as neural networks (and not equations that are immediately interpretable), a proper visualization is crucial to enable easy interpretation of the discovered dynamical model. To do this, we visualize the drift and diffusion functions as fields.

Visualizing $\bof$ is relatively straightforward. Since $\bof : \mathbb R^2 \to \mathbb R^2$ is a vector function, it can be visualized as a vector field---see Figure~\ref{fig:fields-sim} (a) for an example. The arrows indicate the strength and direction of the drift field at each value of $\bm$. The vector field representation makes it easy to identify features equilibrium points (the vectors will have zero length at the equilibrium point) and their stability (near an equilibrium point, the vectors will point towards it if it is stable, and away from it if it is unstable).

Visualizing the diffusion field is more intricate: since $\bg: \mathbb R^2 \to \mathbb R^{2 \times 2}$ is a matrix function, the diffusion field is a tensor field. First, notice that for a given $\bm$, the covariance matrix of $\bm$ is given by $G(\bm) = \bg(\bm) \bg(\bm)^T$. For any value of $\bm$, $G(\bm)$ can be represented by an ellipse centered at $\bm$, with its axes parallel to the eigenvectors of $G(\bm)$ and axis lengths proportional to the eigenvalues of $G(\bm)$. This is a representation of the diffusion tensor field, and gives an overview of the strength and directionality of the diffusive force for each value of $\bm$. For readers familiar with tensor field visualizations, this is a visualization of the $G$-tensor with elliptical glyphs.
   
% \section{Training and architecture details:}

% We carried out all our model training in Tensorflow[cite]. A simple feed-forward neural network architecture with 5 layers was used, each layer having 100 perceptrons. We used elu activation function[cite], which was decided based on trial and error. Training was carried on with batch size of 128. Ten percent of data was set aside for testing. Adamax optimizer module[cite] was used for optimising the network.

% Two networks were initialized, one predicting the mean and other predicting the variance. Both the neural networks were optimised simultaneously using the combined loss function.

% We see great value around developing and using neural netowrk architecture's that can better explain the collective motion phenomenon, looking into equivariant network's can be promising exploring direction's.

\section{Results}
\subsection{Mesoscopic equations for simulation models of collective behaviour}

We first used our approach to discover the mesoscopic dynamics from simulated trajectories, for a class of models where the mesoscopic SDEs are already known---see Section~\ref{sec:datasets}. This serves as a test-bed for the efficacy of the approach. With pairwise as well as ternary interactions, the simulated flocks show a high degree of polarization. However, the mechanism by which order is created is fundamentally different in these two models. 

Figure~\ref{fig:metrics-sim} compares the histogram and autocorrelation of a simulated time series from the discovered models, to that of the input time series used to train the model (also see Table~\ref{comparison}). There is a close match between the input and the generated data (\emph{pairwise:} $W_1 = 0.0686, \trel = 0.2770$, \emph{ternary:} $W_1 = 0.0270, \trel = 0.0437$). We emphasize that this is despite the model not being directly trained to match the histogram or the autocorrelation---suggesting that the discovered models are good approximations for the actual mesoscopic dynamics.

% The training performance of NSMM on the simulated datasets is shown in (FIG). The model was trained for [XXX] epochs, and validation loss falls below [XXX] within the training duration. At the end of the training process, we generated simulated time-series from the trained models. The histogram of the simulated time-series closely matches the histogram of the original time-series (KL-divergence = [XXX]), suggesting that the discovered model is a good approximation is the actual (unknown) generative model.

\begin{figure*}
    \centering
    \includegraphics[width=\textwidth]{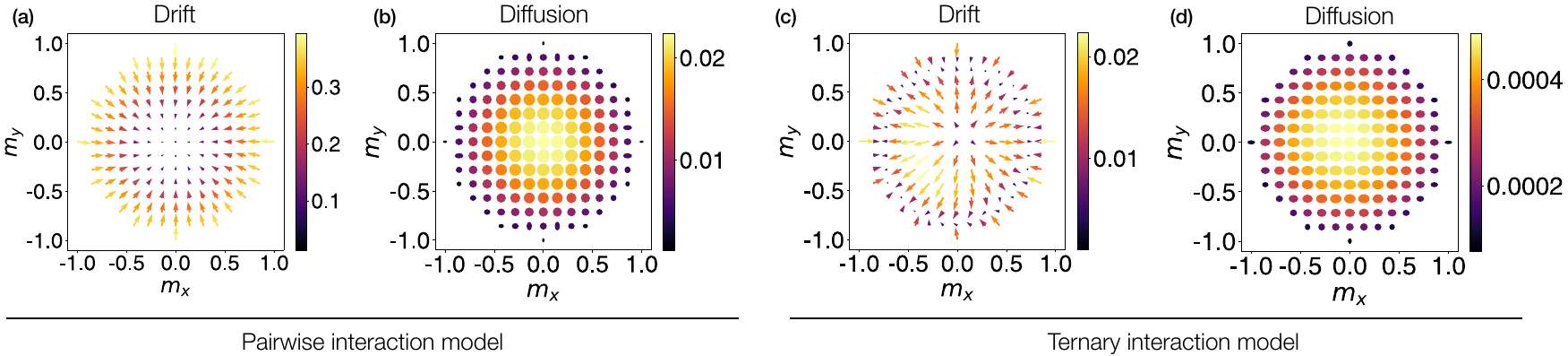}
    \caption{\textbf{Mesoscopic dynamics of simulated fish schools.} The drift and diffusion fields estimated by the neural model agree with the theoretically predicted drift and diffusion fields. (a) Drift field for a simulated school with only pairwise interactions, showing a single attractor at $\bm = 0$. (b) Diffusion field for the pairwise model simulation, showing that the strength of diffusion is maximum at $\bm = 0$ and decreases outwards. (c) Drift field for a simulated school with higher-order (ternary) interactions, showing a ring-shaped attractor for a high value of $|\bm|$. (d) Diffusion field for the ternary model simulation, showing the same pattern of diffusion as in the pairwise model.}
    \label{fig:fields-sim}
\end{figure*}

The estimated drift and diffusion fields are shown in Figure~\ref{fig:fields-sim}. For a model with pairwise interactions, the drift field shows a single attractor at the origin, denoting that the deterministic dynamics has the effect of depolarising the group. However, the strength of diffusion is maximum at the origin and decreases as $\modm$: this means that the stochastic forces push the system away from the $\bm = 0$ stable state, creating order in the system. This counter-intuitive phenomenon of high polarisation arising as a consequence of stochastic effects is called \emph{noise induced order} in literature.

Contrast this with the case of the ternary interaction model, where the drift field shows ring-shaped attractor, at a high value of $\modm$. The diffusion field is similar to the pairwise model, with diffusion being strongest near the origin and decreasing outwards. This means that the order in the ternary interaction model is primarily deterministic.

This demonstrates that our approach is able to distinguish between the fundamentally different dynamics in the two models considered, despite the time series and the histogram of the datasets looking qualitatively similar. Indeed, the estimated models are also in qualitative agreement with theoretical predictions~(Appendix~\ref{sec:analytical}).

\subsection{\label{sec:real-fish} Mesoscopic equations for real-world fish schools}

\begin{figure}
    \centering
    \includegraphics[width=\linewidth]{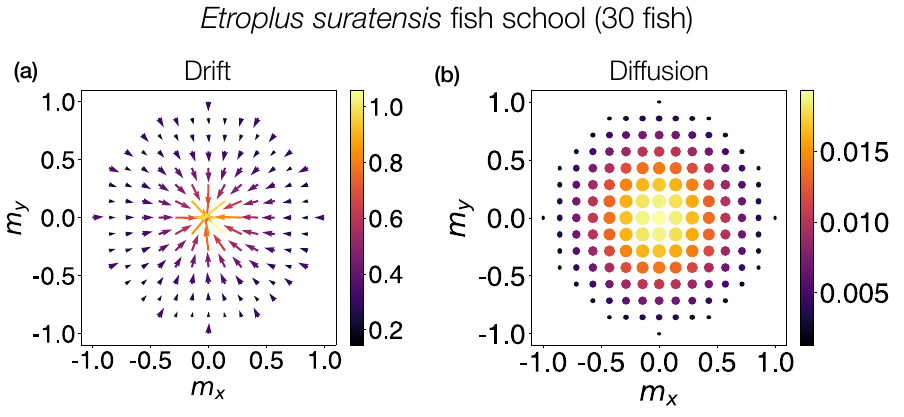}
    \caption{\textbf{Mesoscopic dynamics of a real-world school of 30 \emph{Etroplus suratensis} fish.} The observed drift and diffusion fields show evidence for noise-induced order. (a) Drift field for the \emph{Etroplus school}, showing a single attractor at $\bm = 0$. (b) Diffusion field for school, showing that the strength of diffusion is maximum at $\bm = 0$ and decreases outwards.}
    \label{fig:fields-real}
\end{figure}

We now move on to modelling the stochastic mesoscopic dynamics of a real-world data set of collective motion, based on a recently published dataset on schools of fish~\cite{jhawar2020fish}. Similar to the simulation models, we seek to discover mesoscopic SDEs for the dynamics of the polarization order-parameter $\bm$ of real schools of fish.

% This deviation is interesting, as it suggests deviations in the datasets from the assumptions of the drift-diffusion SDE framework, especially Markovianity and stationarity.
% We hypothesize that the deviation is because of a spurious periodicity in the time series of $|\bm|$---in the experimental trajectory, there are intervals of time where the fish actively follow the boundary of the circular tank, causing a non-stationary oscillation in the time series of $\bm$. This can lead to residual temporal correlations in $|\bm|$. This is a spurious effect arising due to the oscillatory non-stationarity in the data; so we do not expect it to affect the results of the subsequent paragraphs. Indeed, the autocorrelation functions of individual components $m_x$ and $m_y$ of the actual and simulated time series show a good correspondence in the overall decay, barring the oscillations~\textcolor{red}{(APPENDIX)}. 

% The training performance of NSMM on the simulated datasets is shown in (FIG). The model was trained for [XXX] epochs, and validation loss falls below [XXX] within the training duration. The histograms show the comparison between the distributions of the actual and simulated (using the model discovered by NSMM) time-series, and show a close match (KL-divergence = [XXX]).

We observe that the deterministic mean-field dynamics of the school, characterized by the drift field, drives the system towards a disordered state ($\bm = 0$). However, the diffusion is maximum at $\bm = 0$, driving the system away from the stable state and resulting in noise induced order, similar to the model system with pairwise interactions. This surprising phenomenon was first observed in real-world fish schools in~\cite{jhawar2020fish}; our neural approach is able to reproduce this result. This is a non-trivial result: despite the neural networks being able to represent a much larger hypothesis class, the drift and diffusion fields converged to the simple forms shown in Figure~\ref{fig:fields-real}.

We also derived mesoscopic models for fish schools of group sizes 15 and 60. The results are described in Appendix~\ref{sec:groupsizes} and Figure~\ref{fig:groupsizes}, and are qualitatively similar to the results for the 30-fish school. The net strength of diffusion goes down as the school size increases, an observation that matches theoretical predictions (Appendix~\ref{sec:analytical}).

\begin{figure}
    \centering
    \includegraphics[width=\linewidth]{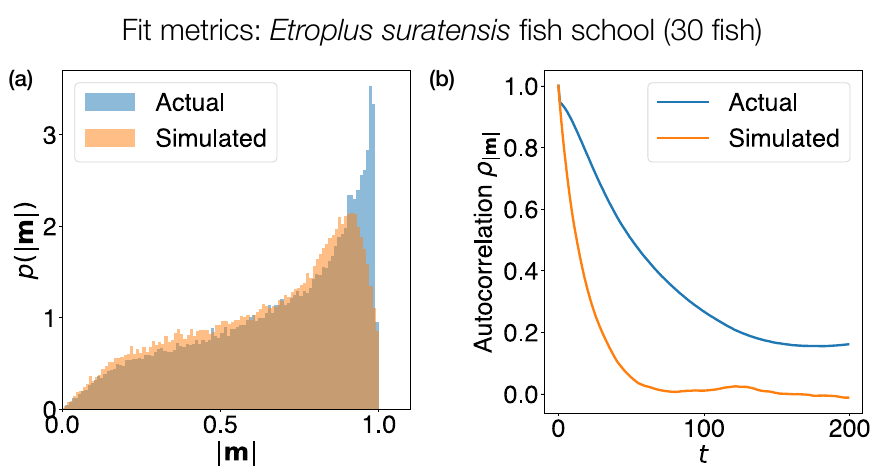}
    \caption{\textbf{Goodness-of-fit metrics for mesoscale models for a real-world fish school.} (a) Comparison of histograms of $|\bm|$ from the actual time series as well as a time series simulated with the discovered neural stochastic model; for the schooling dynamics of a school of 30 fish. (b) Comparison of the autocorrelation functions between the real and simulated time series for the same dataset.}
    \label{fig:metrics-real}
\end{figure}

The fit metrics for the fish schooling dataset (30 fish) is shown in Figure~\ref{fig:metrics-real} (also see Table~\ref{comparison}). Like the agent-based models, there is a good agreement between the histograms of the actual and the simulated $|\bm|$ time series ($W_1 = 0.0823$). However, autocorrelation function shows a deviation from the actual time-series ($\trel = 0.6652$). This deviation is present in other group sizes also (15 and 60 fish). This deviation is because of a spurious periodicity in the time series of $|\bm|$---caused by the fish swimming along the boundary of the tank---which is not captured by the mesoscopic SDE model. The deviation is also present with other SDE estimation techniques (see Table~\ref{comparison}), and is a constraint of the model itself, and not the estimation procedure.

Table~\ref{comparison} summarizes the quantitative performance of the neural effective SDE estimation technique, on the five different datasets considered in this study (2 simulated agent-based models, 3 real-world fish schools of different sizes). For comparison, we use a package PyDaDDy~\cite{nabeel2022pydaddy}, which estimates SDEs from data using classical techniques, \textit{viz.} jump-moment estimation and equation-learning with sparse-regression. While we reemphasize that the focus of a study like this is to obtain interpretable models; overall, the neural SDE estimation approach performs quantitatively better than PyDaddy, in both the Wasserstein metric $W_1$ and relative timescale discrepancy $\trel$.

% \paragraph{Effect of confinement on schooling behavior:} In laboratory studies of collective animal movement, the organisms are usually confined in a bounded arena. It is reasonable to ask if the arena causes boundary effects that meaningfully alters the collective dynamics. To study boundary effects, we study how the mean-field equations change as a function of $r$, the distance from the center of the (circular) tank.

% \textcolor{red}{The following paragraph may change based on the result.} We discovered SDE models of the form in \ref{eq:sde-2}, where $\bf(\bm, r)$ and $\bg(\bm, r)$ are now functions of both $\bm$ and $r$. (FIG) shows the drift and diffusion fields for 3 different values of $r$. We observe that the dynamics remain qualitatively similar across $r$, i.e. the boundary effects do not qualitatively change the stochastic dynamics of schooling.

% \textcolor{red}{[Optional paragraph or non-linear $r$-dependence of "net drift strength", "net diffusion strength etc. Need come up with ways to compute and quantify these.]}

\section{Discussion}

Compared to conventional techniques for SDE identification, neural networks have the advantage of being less restrictive in the hypothesis space they encompass, and hence are more general. It is worth noting that, even with the added flexibility, the neural approach still discovers qualitatively the same model for fish schools as \cite{jhawar2020fish}, lending further credence to the observation of noise-induced order in the original study.

%The specific datasets demonstrated in this study do not fully utilize this flexibility---the inherent rotational and mirror symmetries in the data enable the use of simpler models. 

Indeed, we expect the neural-network based approach to be more powerful---even necessary---in scenarios with more complex movement, which lack these symmetries. For example, organisms could be moving under the influence of an inhomogeneous light or chemical field~\cite{tian2021chemotaxis,puckett2018phototaxis}. In such cases, enforcing user-defined parametric models, or relying on simple hypothesis classes (such as polynomials), may impose strong and often undesirable inductive biases on the SDE discovery process, which the neural-net-based approach avoids.

Further, the likelihood-based estimation procedure is readily generalizable to more complex stochastic dynamics. While such extensions are out of the scope of the current work, it is worth highlighting that the framework is quite general, and can be extended to accommodate jump discontinuities, extrinsic driver variables, non-stationarity, etc., all of which are commonly observed in real-world datasets~\cite{carpenter2020stochastic,salmaso2000factors}.

Finally, we emphasize the need for proper visualization tools to facilitate interpretability of the discovered models. When the goal of model discovery is scientific inquiry (and not merely to obtain a predictive model), tools to interpret and analyze the discovered models become crucial. Our proposed way of visualizing the drift and diffusion fields achieves exactly this, and enables one to easily understand the deterministic phase portrait and the stochasticity landscape of the discovered model.

\begin{table}
\caption{Comparison of model performance with a classical technique for data-driven SDE discovery}
\label{comparison}
\vskip 0.15in
\begin{center}
\begin{small}
\begin{tabular}{lcccr}
\toprule
% \begin{sc}
\textbf{Dataset} & 
\multicolumn{2}{c}{\textbf{Neural SDE}} & 
\multicolumn{2}{c}{\textbf{PyDaddy}} \\
\cmidrule(lr){2-3}\cmidrule(lr){4-5}
                    & $W_1$ & $\trel$ & $W_1$ & $\trel$ \\
% \end{sc}
\midrule
ABM (Pairwise)      & \textbf{0.0686} & \textbf{0.2770} & 0.767 & 0.3020 \\
ABM (Ternary)       & \textbf{0.0270} & \textbf{0.0437} & 0.0283 & 0.1583 \\
Etroplus (15 fish)  & 0.0823 & \textbf{0.6652} & \textbf{0.0818} & 0.7787\\
Etroplus (30 fish)  & 0.0357 & \textbf{0.5895} & \textbf{0.0263} & 0.7144 \\
Etroplus (60 fish)  & \textbf{0.0155} & \textbf{0.2217} & 0.0302 & 0.4409 \\
\bottomrule
\end{tabular}
\end{small}
\end{center}
\vskip -0.1in
\end{table}

\section{Conclusion}

In this study, we utilize a physics-informed, grey-box neural network model to discover mesoscale SDEs for collective movement~\cite{dietrich2021esde,evangelou2022esde}. Using this technique, we discover SDEs for both simulated and real-world datasets of collective movement. With simulated datasets, this approach is able to identify the difference between different models, even when the time series and histograms look qualitatively similar. Furthermore, the SDEs discovered by this approach in each case are are qualitatively similar to the ones predicted by theory. For a real-world dataset of schooling fish, the neural-net approach discovers an SDE that predicts \emph{noise-induced order}, reproducing the result of a recent study~\cite{jhawar2020fish} that used conventional SDE estimation techniques. 

To conclude, we demonstrate the applicability and versatility of neural network models for studying stochastic dynamics of collective animal movement, and of biological systems in general. The combination of a physics-informed, grey-box estimation process along with proper visualization techniques enabled us to discover models that are quantitatively more accurate than traditional approaches (Table~\ref{comparison}), while still remaining readily interpretable. As high-quality datasets of complex phenomena become commonplace in biology and related fields, these techniques---and clear ways to interpret them---become indispensable. 

\section*{Code \& Data Availability}
The fish schooling dataset used in this study is an openly available dataset, available at~\cite{jiteshjhawar_2020_3632470}.

The neural SDE identification code was a modified version of ~\cite{Dietrich2021}. Our modified version of the code is available at \href{https://github.com/utkarshp1161/Neu_sde}{https://github.com/utkarshp1161/Neu\_sde}.

\section*{Acknowledgements}
We thank Vivek Jadhav for his help with the agent-based model simulations, and Somaditya Santra for discussions. P. A. P. acknowledges funding from the Infosys foundation.

\bibliographystyle{icml2023}
\bibliography{references}

\begin{thebibliography}{46}
\providecommand{\natexlab}[1]{#1}
\providecommand{\url}[1]{\texttt{#1}}
\expandafter\ifx\csname urlstyle\endcsname\relax
  \providecommand{\doi}[1]{doi: #1}\else
  \providecommand{\doi}{doi: \begingroup \urlstyle{rm}\Url}\fi

\bibitem[Alert \& Trepat(2020)Alert and Trepat]{alert2020physical}
Alert, R. and Trepat, X.
\newblock Physical models of collective cell migration.
\newblock \emph{Annual Review of Condensed Matter Physics}, 11:\penalty0
  77--101, 2020.

\bibitem[Be’er \& Ariel(2019)Be’er and Ariel]{be2019statistical}
Be’er, A. and Ariel, G.
\newblock A statistical physics view of swarming bacteria.
\newblock \emph{Movement ecology}, 7\penalty0 (1):\penalty0 1--17, 2019.

\bibitem[Biancalani et~al.(2014)Biancalani, Dyson, and
  McKane]{biancalani2014prl}
Biancalani, T., Dyson, L., and McKane, A.~J.
\newblock Noise-induced bistable states and their mean switching time in
  foraging colonies.
\newblock \emph{Physical review letters}, 112\penalty0 (3):\penalty0 038101,
  2014.

\bibitem[Boninsegna et~al.(2018)Boninsegna, Nüske, and
  Clementi]{boninsegna2018sparse}
Boninsegna, L., Nüske, F., and Clementi, C.
\newblock Sparse learning of stochastic dynamic equations.
\newblock \emph{The Journal of Chemical Physics}, 148\penalty0 (24):\penalty0
  241723, June 2018.
\newblock ISSN 0021-9606, 1089-7690.
\newblock \doi{10.1063/1.5018409}.
\newblock URL \url{http://arxiv.org/abs/1712.02432}.
\newblock arXiv: 1712.02432.

\bibitem[Br{\"u}ckner et~al.(2019)Br{\"u}ckner, Fink, Schreiber,
  R{\"o}ttgermann, R{\"a}dler, and Broedersz]{bruckner2019stochastic}
Br{\"u}ckner, D.~B., Fink, A., Schreiber, C., R{\"o}ttgermann, P.~J.,
  R{\"a}dler, J.~O., and Broedersz, C.~P.
\newblock Stochastic nonlinear dynamics of confined cell migration in two-state
  systems.
\newblock \emph{Nature Physics}, 15\penalty0 (6):\penalty0 595--601, 2019.

\bibitem[Buhl et~al.(2006)Buhl, Sumpter, Couzin, Hale, Despland, Miller, and
  Simpson]{buhl2006disorder}
Buhl, J., Sumpter, D.~J., Couzin, I.~D., Hale, J.~J., Despland, E., Miller,
  E.~R., and Simpson, S.~J.
\newblock From disorder to order in marching locusts.
\newblock \emph{Science}, 312\penalty0 (5778):\penalty0 1402--1406, 2006.

\bibitem[Camazine et~al.(2020)Camazine, Deneubourg, Franks, Sneyd, Theraula,
  and Bonabeau]{camazine2020selforg}
Camazine, S., Deneubourg, J.-L., Franks, N.~R., Sneyd, J., Theraula, G., and
  Bonabeau, E.
\newblock Self-organization in biological systems.
\newblock In \emph{Self-Organization in Biological Systems}. Princeton
  university press, 2020.

\bibitem[Carpenter et~al.(2020)Carpenter, Arani, Hanson, Scheffer, Stanley, and
  Van~Nes]{carpenter2020stochastic}
Carpenter, S.~R., Arani, B.~M., Hanson, P.~C., Scheffer, M., Stanley, E.~H.,
  and Van~Nes, E.
\newblock Stochastic dynamics of cyanobacteria in long-term high-frequency
  observations of a eutrophic lake.
\newblock \emph{Limnology and Oceanography Letters}, 5\penalty0 (5):\penalty0
  331--336, 2020.

\bibitem[Chen et~al.(2018{\natexlab{a}})Chen, Rubanova, Bettencourt, and
  Duvenaud]{chen2018neural}
Chen, R.~T., Rubanova, Y., Bettencourt, J., and Duvenaud, D.~K.
\newblock Neural ordinary differential equations.
\newblock \emph{Advances in neural information processing systems}, 31,
  2018{\natexlab{a}}.

\bibitem[Chen et~al.(2018{\natexlab{b}})Chen, Treiber, Kanagaraj, and
  Li]{chen2018social}
Chen, X., Treiber, M., Kanagaraj, V., and Li, H.
\newblock Social force models for pedestrian traffic--state of the art.
\newblock \emph{Transport reviews}, 38\penalty0 (5):\penalty0 625--653,
  2018{\natexlab{b}}.

\bibitem[Clevert et~al.(2015)Clevert, Unterthiner, and
  Hochreiter]{clevert2015elu}
Clevert, D.-A., Unterthiner, T., and Hochreiter, S.
\newblock Fast and accurate deep network learning by exponential linear units
  (elus).
\newblock \emph{arXiv preprint arXiv:1511.07289}, 2015.

\bibitem[Dietrich(2021)]{Dietrich2021}
Dietrich, F.
\newblock {SDE Identification}.
\newblock \url{https://gitlab.com/felix.dietrich/sde-identification}, 2021.

\bibitem[Dietrich et~al.(2021)Dietrich, Makeev, Kevrekidis, Evangelou,
  Bertalan, Reich, and Kevrekidis]{dietrich2021esde}
Dietrich, F., Makeev, A., Kevrekidis, G., Evangelou, N., Bertalan, T., Reich,
  S., and Kevrekidis, I.~G.
\newblock Learning effective stochastic differential equations from microscopic
  simulations: combining stochastic numerics and deep learning.
\newblock \emph{arXiv:2106.09004 [physics]}, 2021.
\newblock URL \url{http://arxiv.org/abs/2106.09004}.
\newblock arXiv: 2106.09004.

\bibitem[Dinet et~al.(2021)Dinet, Michelot, Herrou, and
  Mignot]{dinet2021linking}
Dinet, C., Michelot, A., Herrou, J., and Mignot, T.
\newblock Linking single-cell decisions to collective behaviours in social
  bacteria.
\newblock \emph{Philosophical Transactions of the Royal Society B},
  376\penalty0 (1820):\penalty0 20190755, 2021.

\bibitem[Dyson et~al.(2015)Dyson, Yates, Buhl, and McKane]{dyson2015onset}
Dyson, L., Yates, C.~A., Buhl, J., and McKane, A.~J.
\newblock Onset of collective motion in locusts is captured by a minimal model.
\newblock \emph{Physical Review E}, 92\penalty0 (5):\penalty0 052708, 2015.

\bibitem[Evangelou et~al.(2022)Evangelou, Dietrich, Bello-Rivas, Yeh, Stein,
  Bevan, and Kevekidis]{evangelou2022esde}
Evangelou, N., Dietrich, F., Bello-Rivas, J.~M., Yeh, A., Stein, R., Bevan,
  M.~A., and Kevekidis, I.~G.
\newblock Learning {Effective} {SDEs} from {Brownian} {Dynamics} {Simulations}
  of {Colloidal} {Particles}.
\newblock \emph{arXiv:2205.00286 [cs, math]}, April 2022.
\newblock URL \url{http://arxiv.org/abs/2205.00286}.
\newblock arXiv: 2205.00286.

\bibitem[Gardiner(2009)]{gardiner2009}
Gardiner, C.
\newblock \emph{Stochastic methods}, volume~4.
\newblock springer Berlin, 2009.

\bibitem[Gillespie et~al.(2007)]{gillespie2007stochastic}
Gillespie, D.~T. et~al.
\newblock Stochastic simulation of chemical kinetics.
\newblock \emph{Annual review of physical chemistry}, 58\penalty0 (1):\penalty0
  35--55, 2007.

\bibitem[Gorj{\~a}o \& Meirinhos(2019)Gorj{\~a}o and
  Meirinhos]{gorjao2019kramersmoyal}
Gorj{\~a}o, L.~R. and Meirinhos, F.
\newblock kramersmoyal: Kramers--moyal coefficients for stochastic processes.
\newblock \emph{Journal of Open Source Software}, 4\penalty0 (44):\penalty0
  1693, 2019.

\bibitem[Hasan et~al.(2021)Hasan, Pereira, Farsiu, and
  Tarokh]{hasan2021identifying}
Hasan, A., Pereira, J.~M., Farsiu, S., and Tarokh, V.
\newblock Identifying latent stochastic differential equations.
\newblock \emph{IEEE Transactions on Signal Processing}, 70:\penalty0 89--104,
  2021.

\bibitem[Jhawar \& Guttal(2020)Jhawar and Guttal]{jhawar2020inferring}
Jhawar, J. and Guttal, V.
\newblock Noise-induced effects in collective dynamics and inferring local
  interactions from data.
\newblock \emph{Philosophical Transactions of the Royal Society B},
  375\penalty0 (1807):\penalty0 20190381, 2020.

\bibitem[Jhawar \& Karichannawar(2020)Jhawar and
  Karichannawar]{jiteshjhawar_2020_3632470}
Jhawar, J. and Karichannawar, A.
\newblock {tee-lab/schooling\_fish}, January 2020.
\newblock URL \url{https://doi.org/10.5281/zenodo.3632470}.

\bibitem[Jhawar et~al.(2019)Jhawar, Morris, and Guttal]{jhawar2019bookchapter}
Jhawar, J., Morris, R.~G., and Guttal, V.
\newblock Deriving {Mesoscopic} {Models} of {Collective} {Behavior} for
  {Finite} {Populations}.
\newblock In \emph{Handbook of {Statistics}}, volume~40, pp.\  551--594.
  Elsevier, 2019.
\newblock ISBN 978-0-444-64152-6.
\newblock \doi{10.1016/bs.host.2018.10.002}.

\bibitem[Jhawar et~al.(2020)Jhawar, Morris, Amith-Kumar, Danny~Raj, Rogers,
  Rajendran, and Guttal]{jhawar2020fish}
Jhawar, J., Morris, R.~G., Amith-Kumar, U., Danny~Raj, M., Rogers, T.,
  Rajendran, H., and Guttal, V.
\newblock Noise-induced schooling of fish.
\newblock \emph{Nature Physics}, 16\penalty0 (4):\penalty0 488--493, 2020.

\bibitem[Kloeden et~al.(1992)Kloeden, Platen, Kloeden, and
  Platen]{kloeden1992stochastic}
Kloeden, P.~E., Platen, E., Kloeden, P.~E., and Platen, E.
\newblock \emph{Stochastic differential equations}.
\newblock Springer, 1992.

\bibitem[Li et~al.(2020)Li, Wong, Chen, and Duvenaud]{li2020scalable}
Li, X., Wong, T.-K.~L., Chen, R.~T., and Duvenaud, D.
\newblock Scalable gradients for stochastic differential equations.
\newblock In \emph{International Conference on Artificial Intelligence and
  Statistics}, pp.\  3870--3882. PMLR, 2020.

\bibitem[Liu et~al.(2020)Liu, Long, Wang, Sun, and Dong]{liu2020rode}
Liu, J., Long, Z., Wang, R., Sun, J., and Dong, B.
\newblock Rode-net: learning ordinary differential equations with randomness
  from data.
\newblock \emph{arXiv preprint arXiv:2006.02377}, 2020.

\bibitem[McKane \& Newman(2004)McKane and Newman]{mckane2004stochastic}
McKane, A.~J. and Newman, T.~J.
\newblock Stochastic models in population biology and their deterministic
  analogs.
\newblock \emph{Physical Review E}, 70\penalty0 (4):\penalty0 041902, 2004.

\bibitem[Nabeel et~al.(2022)Nabeel, Karichannavar, Palathingal, Jhawar,
  Danny~Raj, and Guttal]{nabeel2022pydaddy}
Nabeel, A., Karichannavar, A., Palathingal, S., Jhawar, J., Danny~Raj, M., and
  Guttal, V.
\newblock Pydaddy: A python package for discovering stochastic dynamical
  equations from timeseries data.
\newblock \emph{arXiv preprint arXiv:2205.02645}, 2022.

\bibitem[Nielsen et~al.(2000)Nielsen, Madsen, and Young]{nielsen2000parameter}
Nielsen, J.~N., Madsen, H., and Young, P.~C.
\newblock Parameter estimation in stochastic differential equations: an
  overview.
\newblock \emph{Annual Reviews in Control}, 24:\penalty0 83--94, 2000.

\bibitem[O'Leary et~al.(2021)O'Leary, Paulson, and Mesbah]{o2021stochastic}
O'Leary, J., Paulson, J.~A., and Mesbah, A.
\newblock Stochastic physics-informed neural networks (spinn): A
  moment-matching framework for learning hidden physics within stochastic
  differential equations.
\newblock \emph{arXiv preprint arXiv:2109.01621}, 2021.

\bibitem[Puckett et~al.(2018)Puckett, Pokhrel, and
  Giannini]{puckett2018phototaxis}
Puckett, J.~G., Pokhrel, A.~R., and Giannini, J.~A.
\newblock Collective gradient sensing in fish schools.
\newblock \emph{Scientific reports}, 8\penalty0 (1):\penalty0 1--11, 2018.

\bibitem[Ramaswamy(2010)]{ramaswamy2010mechanics}
Ramaswamy, S.
\newblock The mechanics and statistics of active matter.
\newblock \emph{Annual Review of Condensed Matter Physics}, 1\penalty0
  (1):\penalty0 323--345, 2010.
\newblock \doi{10.1146/annurev-conmatphys-070909-104101}.
\newblock URL \url{https://doi.org/10.1146/annurev-conmatphys-070909-104101}.

\bibitem[Ramaswamy(2017)]{ramaswamy2017active}
Ramaswamy, S.
\newblock Active matter.
\newblock \emph{Journal of Statistical Mechanics: Theory and Experiment},
  2017\penalty0 (5):\penalty0 054002, 2017.

\bibitem[Ramdas et~al.(2017)Ramdas, Garc{\'\i}a~Trillos, and
  Cuturi]{ramdas2017wasserstein}
Ramdas, A., Garc{\'\i}a~Trillos, N., and Cuturi, M.
\newblock On wasserstein two-sample testing and related families of
  nonparametric tests.
\newblock \emph{Entropy}, 19\penalty0 (2):\penalty0 47, 2017.

\bibitem[R{\o}rth(2009)]{rorth2009cellmigration}
R{\o}rth, P.
\newblock Collective cell migration.
\newblock \emph{Annual Review of Cell and Developmental Biology}, 25:\penalty0
  407--429, 2009.

\bibitem[Salmaso(2000)]{salmaso2000factors}
Salmaso, N.
\newblock Factors affecting the seasonality and distribution of cyanobacteria
  and chlorophytes: a case study from the large lakes south of the alps, with
  special reference to lake garda.
\newblock \emph{Hydrobiologia}, 438\penalty0 (1):\penalty0 43--63, 2000.

\bibitem[Song et~al.(2020)Song, Sohl-Dickstein, Kingma, Kumar, Ermon, and
  Poole]{song2020score}
Song, Y., Sohl-Dickstein, J., Kingma, D.~P., Kumar, A., Ermon, S., and Poole,
  B.
\newblock Score-based generative modeling through stochastic differential
  equations.
\newblock \emph{arXiv preprint arXiv:2011.13456}, 2020.

\bibitem[Sumpter(2010)]{sumpter2010collective}
Sumpter, D.~J.
\newblock Collective animal behavior.
\newblock In \emph{Collective Animal Behavior}. Princeton University Press,
  2010.

\bibitem[Tabar(2019)]{tabar2019book}
Tabar, R.
\newblock \emph{Analysis and data-based reconstruction of complex nonlinear
  dynamical systems}, volume 730.
\newblock Springer, 2019.

\bibitem[Tian et~al.(2021)Tian, Zhang, Zhang, and Yuan]{tian2021chemotaxis}
Tian, M., Zhang, C., Zhang, R., and Yuan, J.
\newblock Collective motion enhances chemotaxis in a two-dimensional bacterial
  swarm.
\newblock \emph{Biophysical journal}, 120\penalty0 (9):\penalty0 1615--1624,
  2021.

\bibitem[Van~Kampen(1992)]{van1992stochastic}
Van~Kampen, N.~G.
\newblock \emph{Stochastic processes in physics and chemistry}, volume~1.
\newblock Elsevier, 1992.

\bibitem[Vicsek \& Zafeiris(2012)Vicsek and Zafeiris]{vicsek2012collective}
Vicsek, T. and Zafeiris, A.
\newblock Collective motion.
\newblock \emph{Physics reports}, 517\penalty0 (3-4):\penalty0 71--140, 2012.

\bibitem[Vicsek et~al.(1995)Vicsek, Czirók, Ben-Jacob, Cohen, and
  Shochet]{vicsek1995novel}
Vicsek, T., Czirók, A., Ben-Jacob, E., Cohen, I., and Shochet, O.
\newblock Novel {Type} of {Phase} {Transition} in a {System} of {Self}-{Driven}
  {Particles}.
\newblock \emph{Physical Review Letters}, 75\penalty0 (6):\penalty0 1226--1229,
  August 1995.
\newblock ISSN 0031-9007, 1079-7114.
\newblock \doi{10.1103/PhysRevLett.75.1226}.
\newblock URL \url{https://link.aps.org/doi/10.1103/PhysRevLett.75.1226}.

\bibitem[Wang et~al.(2022)Wang, Fang, Jin, Ma, He, Dai, Yue, Cheng, Zhang, Pu,
  et~al.]{wang2022data}
Wang, Y., Fang, H., Jin, J., Ma, G., He, X., Dai, X., Yue, Z., Cheng, C.,
  Zhang, H.-T., Pu, D., et~al.
\newblock Data-driven discovery of stochastic differential equations.
\newblock \emph{Engineering}, 2022.

\bibitem[Yates et~al.(2009)Yates, Erban, Escudero, Couzin, Buhl, Kevrekidis,
  Maini, and Sumpter]{yates2009locust}
Yates, C.~A., Erban, R., Escudero, C., Couzin, I.~D., Buhl, J., Kevrekidis,
  I.~G., Maini, P.~K., and Sumpter, D. J.~T.
\newblock Inherent noise can facilitate coherence in collective swarm motion.
\newblock \emph{Proceedings of the National Academy of Sciences}, 106\penalty0
  (14):\penalty0 5464--5469, April 2009.
\newblock ISSN 0027-8424, 1091-6490.
\newblock \doi{10.1073/pnas.0811195106}.
\newblock URL \url{https://pnas.org/doi/full/10.1073/pnas.0811195106}.

\end{thebibliography}

\newpage
\appendix
\onecolumn

\section{\label{sec:analytical} Analytically derived mesoscopic SDEs for agent-based models}

\begin{figure*}
    \centering
    \includegraphics[width=\textwidth]{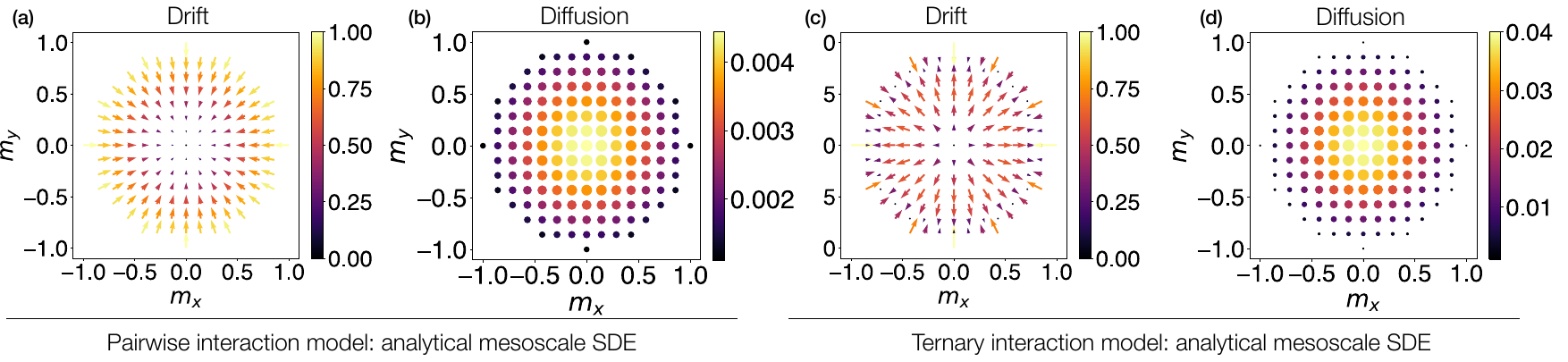}
    \caption{\textbf{Analytically derived mesoscale dynamics of agent based models.} (a) Analytically derived drift field for a model with only pairwise interactions, showing a single attractor at $\bm = 0$. (b) Analytically derived diffusion field for the pairwise model, showing that the strength of diffusion is maximum at $\bm = 0$ and decreases outwards. (c) Analytically derived drift field for a model with higher-order (ternary) interactions, showing a ring-shaped attractor for a high value of $|\bm|$. (d) Analytically derived diffusion field for the ternary model simulation, showing the same pattern of diffusion as in the pairwise model.}
    \label{fig:fields-analytical}
\end{figure*}

The agent-based models of collective behaviour used in this study have been well-studied, and mesoscopic SDEs for these models have been analytically derived~\cite{jhawar2019bookchapter, jhawar2020fish}.

Recap the model description:
\begin{itemize}
    \item An agent $i$ can spontaneously turn and choose a new direction, i.e. $\theta_i(t) \rightarrow \eta$ for $\eta \sim \mathsf{Unif}[-\pi, \pi)$. The spontaneous turns happen as a Poisson process with a rate $r_1$.
    \item An agent $i$ can copy the direction of one (pairwise interaction) or the average direction of two (ternary interaction) other agents, chosen randomly from the group. The copy interactions can happen at rates $r_2$ and $r_3$ respectively.
\end{itemize}

For the pairwise interaction model, the mesoscale SDE can be analytically derived to be of the following form:

\begin{align}
    \frac{d \bm}{dt} &=
    -r_1 \bm +
    \sqrt{ \frac{r_1 + r_2\left(1 - \modm^2\right)}{N}} I \cdot \boldeta(t) \label{eq:analytical-pairwise}
\end{align}

where $N$ is the total number of agents in the group.

Similarly, for the ternary interaction model, the derived mesoscale SDE has the form:

\begin{align}
    \frac{d \bm}{dt} &=
    -r_1 \bm + r_3 \left(1 - \modm^2 \right) \bm +
    \sqrt{ \frac{r_1 + (r_2 + r_3)\left(1 - \modm^2\right)}{N}} I \cdot \boldeta(t) \label{eq:analytical-ternary}
\end{align}

The diffusion term is a diagonal matrix, and has a $1 - \modm^2$ term. This means that the noise is maximum at $\modm = 0$ and decreases with increasing $\modm$. The drift term is linear in $\bm$ for the pairwise interaction model, and the deterministic stable equilibrium for this system is at $\modm = 0$. Therefore, the order in the pairwise interaction model is noise-induced, and arises solely due to the high strength of diffusion near $\modm = 0$. On the other hand, for the ternary interaction model, drift term is cubic in $\bm$, and has a $\left(1 - \modm^2\right)\bm$ term. This contributes to a ring-shaped attractor near $\modm = 1$, i.e. the order in the ternary interaction model originates from a combination of both deterministic and noise-induced effects.

The drift and diffusion fields generated from equations \ref{eq:analytical-pairwise} and \ref{eq:analytical-ternary} are shown in Figure~\ref{fig:fields-analytical}. There is a close correspondence between these plots and the ones in Figure~\ref{fig:fields-sim}, suggesting that the neural effective SDE approach is able to correctly infer the underlying mesoscale SDEs.

\section{\label{sec:groupsizes} Mesoscopic models for fish schools of different group sizes}

\begin{figure}[t]
    \centering
    \includegraphics[width=\textwidth]{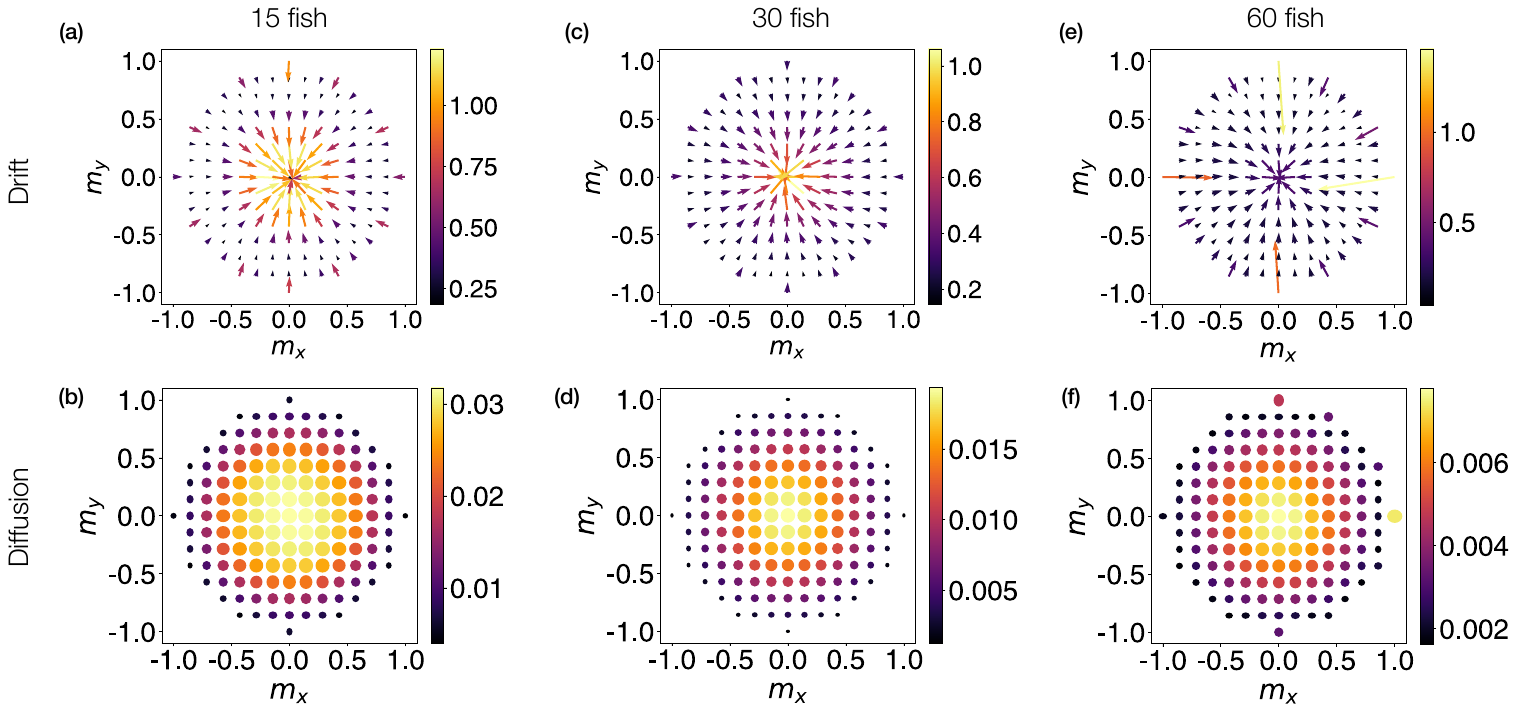}
    \caption{\textbf{Estimated drift and diffusion fields for polarization dynamics of fish schools of different group sizes.} (a, b) Drift and diffusion fields for $\bm$ for a 15-fish school (c, d) Drift and diffusion fields for $\bm$ for a 15-fish school, same as in Figure~\ref{fig:fields-real}. (e, f) Drift and diffusion fields for $\bm$ for a 60-fish school.}
    \label{fig:groupsizes}
\end{figure}

We repeated the analysis in Section~\ref{sec:real-fish} on two other group sizes, \emph{viz.} $N=15$ and $N=60$. The fit metrics and drift/diffusion fields are shown in Figure.~\ref{fig:groupsizes}. Overall, the fields look qualitatively similar across different group sizes. However, the net strength of diffusion decreases with increasing group size (notice the range of the color axis). This is in accordance with theoretical predictions: in theory, we expect the diffusion term $\bg$ to scale as $1/\sqrt{N}$ (see Equation~\ref{eq:analytical-pairwise}). 

\end{document}